\documentclass[a4paper,twoside,english,amscd,amsmath,amssymb,verbatim]{article}
\usepackage[T1]{fontenc}
\usepackage[latin9]{inputenc}
\usepackage{babel}
\usepackage{amsmath}
\usepackage{esint}
\usepackage[unicode=true,pdfusetitle,
 bookmarks=true,bookmarksnumbered=false,bookmarksopen=false,
 breaklinks=false,pdfborder={0 0 1},backref=false,colorlinks=false]
 {hyperref}

\makeatletter

\pdfpageheight\paperheight
\pdfpagewidth\paperwidth

\makeatother

\begin{document}

\title{Some Theoretical Properties of a Network of Discretely Firing Neurons\thanks{This paper was submitted to Special Issue of Neurocomputing on Theoretical
Analysis of Real-Valued Function Classes on 19 January 1998. It was
not accepted for publication, but it underpins several subsequently
published papers.}}

\author{Stephen Luttrell}

\maketitle
\textbf{Abstract:} The problem of optimising a network of discretely
firing neurons is addressed. An objective function is introduced which
measures the average number of bits that are needed for the network
to encode its state. When this is minimised, it is shown that this
leads to a number of results, such as topographic mappings, piecewise
linear dependence on the input of the probability of a neuron firing,
and factorial encoder networks.

\section{Introduction}

In this paper the problem of optimising the firing characteristics
of a network of discretely firing neurons will be considered. The
approach adopted will not be based on any particular model of how
real neurons operate, but will focus on theoretically analysing some
of the information processing capabilities of a layered network of
units (which happen to be called neurons). Ideal network behaviour
is derived by choosing the ideal neural properties that minimise an
information theoretic objective function which specifies the number
of bits required by the network to encode the state of its layers.
This is done in preference to assuming a highly specific neural behaviour
at the outset, followed by optimisation of a few remaining parameters
such as weight and bias values.

Why use an objective function in the first place? An objective function
is a very convenient starting point (a set of ``axioms'', as it
were), from which everything else can, in principle, be derived (as
``theorems'', as it were). An objective function has the same status
as a model, which may be falsified should some counterevidence be
discovered. The objective function used in this paper is the simplest
that is consistent with predicting a number of non-trivial results,
such as topographic mappings, and factorial encoders (which are discussed
in this paper). However, it does not include any temporal information,
nor any biological plausibility constraints (other than the fact that
the network is assumed to be layered). More complicated objective
functions will be the subject of future publications.

In section \ref{sec:Theory} an objective function is introduced,
and its connection with discretely firing neural networks is derived.
In section \ref{sec:Examples} some examples are presented which show
how this theory of discretely firing neural networks leads to some
non-trivial results.

\section{Theory\label{sec:Theory}}

In this section a theory of discretely firing neural networks is developed.
Section \ref{sub:Objective-Function-for} introduces the objective
function for optimising an encoder, and section \ref{sub:Application-to-Neural}
shows how this can be applied to the problem of optimising a discretely
firing neural network.

\subsection{Objective Function for Optimal Coding\label{sub:Objective-Function-for}}

The inspiration for the approach that is used here is the minimum
description length (MDL) method \cite{Rissanen1978}. In this paper,
a training set vector (which is unlabelled) will be denoted as $\mathbf{x}$,
a vector of statistics which are stochastically derived from $\mathbf{x}$
will be denoted as $\mathbf{y}$, and their joint probability density
function (PDF) will be denoted as $\Pr(\mathbf{x},\mathbf{y})$. The
problem is to learn the functional form of $\Pr(\mathbf{x},\mathbf{y})$,
so that vectors $(\mathbf{x},\mathbf{y})$ sampled from $\Pr(\mathbf{x},\mathbf{y})$
can be encoded using the minimum number of bits on average. It is
unconventional to consider the problem of encoding $(\mathbf{x},\mathbf{y})$,
rather than $\mathbf{x}$ alone, but it turns out that this leads
to many useful results.

Thus $\Pr(\mathbf{x},\mathbf{y})$ is approximated by a learnt model
$Q(\mathbf{x},\mathbf{y})$, in which case the average number of bits
required to encode an $(\mathbf{x},\mathbf{y})$ sampled from the
PDF $\Pr(\mathbf{x},\mathbf{y})$ is given by the objective function
$D$, which is defined as 
\begin{equation}
D\equiv-\int d\mathbf{x}\sum_{\mathbf{y}}\Pr(\mathbf{x},\mathbf{y})\,\log Q(\mathbf{x},\mathbf{y})\label{eq:Objective}
\end{equation}
Now split $D$ into two contributions by using $\Pr(\mathbf{x},\mathbf{y})=\Pr(\mathbf{x})\,\Pr(\mathbf{y}|\mathbf{x})$
and $Q(\mathbf{x},\mathbf{y})=Q(\mathbf{x})\,Q(\mathbf{y}|\mathbf{x})$.
\begin{equation}
D=-\int d\mathbf{x}\,\Pr(\mathbf{x})\sum_{\mathbf{y}}\Pr(\mathbf{y}|\mathbf{x})\,\log Q(\mathbf{x}|\mathbf{y})-\sum_{\mathbf{y}}\Pr(\mathbf{y})\,\log Q(\mathbf{y})\label{eq:ObjectiveSplit}
\end{equation}
The first term is the cost (i.e. the average number of bits), averaged
over all possible values of $\mathbf{y}$, of encoding an $\mathbf{x}$
sampled from $\Pr(\mathbf{x}|\mathbf{y})$ using the model $Q(\mathbf{x}|\mathbf{y})$.
This interpretation uses that $\Pr(\mathbf{x})\,\Pr(\mathbf{y}|\mathbf{x})=\Pr(\mathbf{y})\,\Pr(\mathbf{x}|\mathbf{y})$.
The second term is the cost of encoding a $\mathbf{y}$ sampled from
$\Pr(\mathbf{y})$ using the model $Q(\mathbf{y})$. Together these
two terms correspond to encoding $\mathbf{y}$ (the second term),
then encoding $\mathbf{x}$ given that $\mathbf{y}$ is known.

The model $Q(\mathbf{x},\mathbf{y})$ may be optimised so that it
minimises $D$, and thus leads to the minimum cost of encoding $(\mathbf{x},\mathbf{y})$
sampled from $\Pr(\mathbf{x},\mathbf{y})$. Ideally $Q(\mathbf{x},\mathbf{y})=\Pr(\mathbf{x},\mathbf{y})$,
but in practice this is not possible because insufficient information
is available to determine $\Pr(\mathbf{x},\mathbf{y})$ exactly (i.e.
the training set does not contain an infinite number of $(\mathbf{x},\mathbf{y})$
vectors). It is therefore necessary to introduce a parametric model
$Q(\mathbf{x},\mathbf{y})$, and to choose the values of the parameters
so that $D$ is minimised. If the number of parameters is small enough,
and the training set is large enough, then the parameter values can
be accurately determined.

A further simplification may be made if $\mathbf{y}$ can occupy much
fewer states than $\mathbf{x}$ (given $\mathbf{y}$) can, because
then the cost of encoding $\mathbf{y}$ is much less than the cost
of encoding $\mathbf{x}$ (given $\mathbf{y}$) (i.e. the second and
first terms in equation \ref{eq:ObjectiveSplit}, respectively). In
this case, it is a good approximation to retain only the first term
in equation \ref{eq:ObjectiveSplit}. This approximation becomes exact
if $Q(\mathbf{y})$ assigns equal probability to all states $\mathbf{y}$,
because then the third term is a constant. The reason for defining
the objective function $D$ as in equation \ref{eq:Objective}, rather
than defining it to be the first term of equation \ref{eq:ObjectiveSplit},
is because equation \ref{eq:Objective} may be readily generalised
to more complex systems, such as $(\mathbf{x},\mathbf{y},\mathbf{z})$
in which $\Pr(\mathbf{x},\mathbf{y},\mathbf{z})=\Pr(\mathbf{x})\Pr(\mathbf{y}|\mathbf{x})\Pr(\mathbf{z}|\mathbf{y})$,
and so on. An example of this is given in section \ref{sub:Topographic-Mapping-Neural}. 

It is possible to relate the minimisation of $D$ to the maximisation
of the mutual information $I$ between $\mathbf{x}$ and $\mathbf{y}$.
If the cost of encoding an $\mathbf{x}$ sampled from $\Pr(\mathbf{x})$
using the model $Q(\mathbf{x})$ (i.e. $-\int d\mathbf{x}\,\Pr(\mathbf{x})\,\log Q(\mathbf{x})$)
and the cost of encoding a $\mathbf{y}$ sampled from $\Pr(\mathbf{y})$
using the model $Q(\mathbf{y})$ (i.e. $-\sum_{\mathbf{y}}\Pr(\mathbf{y})\,\log Q(\mathbf{y})$)
are both subtracted from $D$, then the result is $-\int d\mathbf{x}\,\sum_{\mathbf{y}}\Pr(\mathbf{x},\mathbf{y})\,\log\left(\frac{Q(\mathbf{x},\mathbf{y})}{Q(\mathbf{x})\,Q(\mathbf{y})}\right)$.
When $Q(\mathbf{x},\mathbf{y})\longrightarrow\Pr(\mathbf{x},\mathbf{y})$
this reduces to (minus) the mutual information $I$ between $\mathbf{x}$
and $\mathbf{y}$. Thus, if the cost of encoding the correlations
between $\mathbf{x}$ and $\mathbf{y}$ is much greater than the cost
of separately encoding $\mathbf{x}$ and $\mathbf{y}$ (i.e. the $\log\left(Q(\mathbf{x})\,Q(\mathbf{y})\right)$
term can be ignored in $I$), then $D$-minimisation approximates
$I$-maximisation, which is another commonly used objective function.

\subsection{Application to Neural Networks\label{sub:Application-to-Neural}}

In order to apply the above coding theory results to a 2-layer discretely
firing neural network, it is necessary to interpret $\mathbf{x}$
as a pattern of activity in the input layer, and $\mathbf{y}$ as
the vector of locations in the output layer of a finite number of
firing events. The objective function $D$ is then the cost of using
the model $Q(\mathbf{x},\mathbf{y})$ of the network behaviour to
encode the state $(\mathbf{x},\mathbf{y})$ of the neural network
(i.e. the input pattern and the location of the firing events), which
is sampled from the $\Pr(\mathbf{x},\mathbf{y})$ that describes the
true network behaviour. For instance, a second neural network can
be used solely for computing the model $Q(\mathbf{x},\mathbf{y})$,
which is then used to encode the state $(\mathbf{x},\mathbf{y})$
of the above first neural network. Note that no temporal information
is included in this analysis, so the input and output of the network
is a static $(\mathbf{x},\mathbf{y})$ vector containing no time variables.

These two neural networks can be combined into a single hybrid network,
in which the machinery for computing the model $Q(\mathbf{x},\mathbf{y})$
is interleaved with the neural network, whose true behavior is described
by $\Pr(\mathbf{x},\mathbf{y})$. The notation of equation \ref{eq:ObjectiveSplit}
can now be expressed in more neural terms, where $\Pr(\mathbf{y}|\mathbf{x})$
is then a recognition model (i.e. bottom-up) and $Q(\mathbf{x}|\mathbf{y})$
is then a generative model (i.e. top-down), both of which live inside
the same neural network. This is an unsupervised neural network, because
it is trained with examples of only $\mathbf{x}$-vectors, and the
network uses its $\Pr(\mathbf{y}|\mathbf{x})$ to stochastically generate
a $\mathbf{y}$ from each $\mathbf{x}$.

Now introduce a Gaussian parametric model $Q(\mathbf{x}|\mathbf{y})$
\begin{equation}
Q(\mathbf{x}|\mathbf{y})=\frac{1}{\left(\sqrt{2\pi}\sigma\right)^{\mathrm{dim}\,\mathbf{x}}}\,\,\exp\left(-\frac{\left\Vert \mathbf{x}-\mathbf{x}^{\prime}(\mathbf{y})\right\Vert ^{2}}{2\sigma^{2}}\right)
\end{equation}
where $\mathbf{x}^{\prime}(\mathbf{y})$ is the centroid of the Gaussian
(given $\mathbf{y}$), $\sigma$ is the standard deviation of the
Gaussian. Also define a soft vector quantiser (VQ) objective function
$D_{\mathrm{VQ}}$ as
\begin{equation}
D_{\mathrm{VQ}}\equiv2\int d\mathbf{x}\,\Pr\left(\mathbf{x}\right)\sum_{\mathbf{y}}\Pr(\mathbf{y}|\mathbf{x})\,\left\Vert \mathbf{x}-\mathbf{x}^{\prime}(\mathbf{y})\right\Vert ^{2}
\end{equation}
which is (twice) the average Euclidean reconstruction error that results
when $\mathbf{x}$ is probabilistically encoded as $\mathbf{y}$ and
then deterministically reconstructed as $\mathbf{x}^{\prime}(\mathbf{y})$.
These definitions of $Q(\mathbf{x}|\mathbf{y})$ and $D_{\mathrm{VQ}}$
allow $D$ to be written as 
\begin{equation}
D=\frac{1}{4\sigma^{2}}\,D_{\mathrm{VQ}}-\log\left(\sqrt{2\pi}\sigma\right)\,\dim\mathbf{x}-\sum_{\mathbf{y}}\Pr(\mathbf{y})\,\log Q(\mathbf{y})
\end{equation}
where the second term is constant, and the third term may be ignored
if $\mathbf{y}$ can occupy much fewer states than $\mathbf{x}$ (given
$\mathbf{y}$) can. The conditions under which the third term can
be ignored are satisfed in a neural network, because $\mathbf{x}$
is an activity pattern, and $\mathbf{y}$ as the vector of locations
of a finite number of firing events. 

The first term of $D$ is proportional to $D_{\mathrm{VQ}}$, whose
properties may be investigated using the techniques in \cite{Luttrell1997}.
Assume that there are $n$ firing events, so that $\mathbf{y}=\left(y_{1},y_{2},\cdots,y_{n}\right)$,
then the marginal probabilities of the symmetric part $S\left[\Pr(\mathbf{y}|\mathbf{x})\right]$
of $\Pr(\mathbf{y}|\mathbf{x})$ under interchange of its $\left(y_{1},y_{2},\cdots,y_{n}\right)$
arguments are given by 
\begin{eqnarray}
\Pr(y_{1}|\mathbf{x}) & \equiv & \sum_{y_{2},\cdots,y_{n}=1}^{M}S\left[\Pr(y_{1},y_{2},\cdots,y_{n}|\mathbf{x})\right]\nonumber \\
\Pr(y_{1},y_{2}|\mathbf{x}) & \equiv & \sum_{y_{3},\cdots,y_{n}=1}^{M}S\left[\Pr(y_{1},y_{2},\cdots,y_{n}|\mathbf{x})\right]
\end{eqnarray}
where $\Pr(y_{1}|\mathbf{x})$ may be interpreted as the probability
that the next firing event occurs on neuron $y$ (given $\mathbf{x}$),
Also define 2 useful integrals, $D_{1}$ and $D_{2}$, as 
\begin{eqnarray}
D_{1} & \equiv & \frac{2}{n}\int d\mathbf{x}\,\Pr(\mathbf{x})\sum_{y=1}^{M}\Pr(y|\mathbf{x})\,\left\Vert \mathbf{x}-\mathbf{x}^{\prime}(y)\right\Vert ^{2}\nonumber \\
D_{2} & \equiv & \frac{2(n-1)}{n}\int d\mathbf{x}\,\Pr(\mathbf{x})\sum_{y_{1},y_{2}=1}^{M}\Pr(y_{1},y_{2}|\mathbf{x})\,\left(\mathbf{x}-\mathbf{x}^{\prime}(y_{1})\right)\cdot\left(\mathbf{x}-\mathbf{x}^{\prime}(y_{2})\right)\label{eq:CorrelatedD1D2}
\end{eqnarray}
where $\mathbf{x}^{\prime}(y)$ is any vector function of $y$ (i.e.
not necessarily related to $\mathbf{x}^{\prime}(\mathbf{y})$), to
yield the following upper bound on $D_{\mathrm{VQ}}$
\begin{equation}
D_{\mathrm{VQ}}\leq D_{1}+D_{2}\label{eq:ObjectiveD1D2}
\end{equation}
where $D_{1}$ is non-negative but $D_{2}$ can have either sign,
and the inequality reduces to an equality in the case $n=1$. Thus
far nothing specific has been assumed about $\Pr(\mathbf{y}|\mathbf{x})$,
other than the fact that it contains no temporal information, so the
upper bound on $D_{\mathrm{VQ}}$ applies whatever the form of $\Pr(\mathbf{y}|\mathbf{x})$. 

If the firing events occur independently of each other (given $\mathbf{x}$),
then $\Pr(y_{1},y_{2}|\mathbf{x})=\Pr(y_{1}|\mathbf{x})\,\Pr(y_{2}|\mathbf{x})$,
which allows $D_{2}$ to be redefined as 
\begin{equation}
D_{2}\equiv\frac{2(n-1)}{n}\int d\mathbf{x}\,\Pr(\mathbf{x})\,\left\Vert \mathbf{x}-\sum_{y=1}^{M}\Pr(y|\mathbf{x})\,\mathbf{x}^{\prime}(y)\right\Vert ^{2}\label{eq:IndependentD2}
\end{equation}
where $D_{2}$ is non-negative.

In summary, the assumptions which have been made in order to obtain
the upper bound on $D_{\mathrm{VQ}}$ in equation \ref{eq:ObjectiveD1D2}
with the definition of $D_{1}$ as given in equation \ref{eq:CorrelatedD1D2}
and $D_{2}$ as given in equation \ref{eq:IndependentD2} are: no
temporal information is included in the network state vector $(\mathbf{x},\mathbf{y})$,
$\mathbf{y}$ can occupy much fewer states than $\mathbf{x}$ (given
$\mathbf{y}$) can, and firing events occur independently of each
other (given $\mathbf{x}$). In reality, there is always temporal
information available, and the firing events are correlated with each
other, so a more realistic objective function could be constructed.
However, it is worthwhile to consider the consequences of equation
\ref{eq:ObjectiveD1D2}, because it turns out that it leads to many
non-trivial results. 

The upper bound on $D_{\mathrm{VQ}}$ may be minimised with respect
to all free parameters in order to obtain a least upper bound. In
the case of independent firing events, the free parameters are the
$\mathbf{x}^{\prime}(y)$ and the $\Pr(y|\mathbf{x})$. These two
types of parameters cannot be independently optimised, because they
correspond to the generative and recognition models implicit in the
neural network, respectively. 

A gradient descent algorithm for optimising the parameter values may
readily be obtained by differentiating $D_{1}$ and $D_{2}$ with
respect to $\mathbf{x}^{\prime}(y)$ and $\Pr(y|\mathbf{x})$. Given
the freedom to explore the entire space of functions $\Pr(y|\mathbf{x})$,
the optimum neural firing behaviour (given $\mathbf{x}$) can in principle
be determined, and in certain simple cases this can be determined
by inspection. If this option is not available, such as would be the
case if biological contraints restricted the allowed functional form
of $\Pr(y|\mathbf{x})$, then a limited search of the entire space
of functions $\Pr(y|\mathbf{x})$ can be made by invoking parametric
model of the neural firing behaviour (given $\mathbf{x}$).

\section{Examples\label{sec:Examples}}

In this section several examples are presented which illustrate the
use of $D_{1}+D_{2}$ in the optimisation of discretely firing neural
networks. In section \ref{sub:Topographic-Mapping-Neural} a topographic
mapping network is derived from $D_{1}$ alone, in section \ref{sub:Piecewise-Linear-Probability}
$\Pr(y|\mathbf{x})$ that minimises $D_{1}+D_{2}$ is shown to be
piecewise linear, and a solved example is presented. Finally, in section
\ref{sub:Factorial-Encoder-Network} a more detailed worked example
is presented, which demonstrates how a factorial encoder emerges when
$D_{1}+D_{2}$ is minimised.

\subsection{Topographic Mapping Neural Network\label{sub:Topographic-Mapping-Neural}}

When an appropriate from of $V_{\mathrm{VQ}}$ is considered, it can
be seen that it leads to a network that is closely related to Kohonen's
topographic mapping network \cite{Kohonen1989}. 

The derivation of a topographic mapping network that was given in
\cite{Luttrell1989b} will now be recast in the framework of section
\ref{sub:Application-to-Neural}. Thus, consider the objective function
for a 3-layer network $(\mathbf{x},y,z)$, in which (compare equation
\ref{eq:Objective}) 
\begin{equation}
D=-\int\sum_{z}d\mathbf{x}\,\Pr(\mathbf{x},z)\,\log Q(\mathbf{x},z)
\end{equation}
where the cost of encoding $y$ has been ignored, so that effectively
only a 2-layer network $(\mathbf{x},z)$ is visible, and $D_{\mathrm{VQ}}$
is given by 
\begin{equation}
D_{\mathrm{VQ}}=2\int d\mathbf{x}\,\Pr(\mathbf{x})\sum_{z=1}^{M_{z}}\Pr(z|\mathbf{x})\,\left\Vert \mathbf{x}-\mathbf{x}^{\prime}(z)\right\Vert ^{2}\label{eq:ObjectiveLayer12}
\end{equation}
This expression for $D_{\mathrm{VQ}}$ explicitly involves $(\mathbf{x},z)$,
but it may be manipulated into a form that explicitly involves $(\mathbf{x},y)$.
In order to make simplify this calculation, $D_{\mathrm{VQ}}$ will
be replaced by the equivalent objective function 
\begin{equation}
D_{\mathrm{VQ}}=\int d\mathbf{x}\,\Pr(\mathbf{x})\,\sum_{z=1}^{M_{z}}\Pr(z|\mathbf{x})\int d\mathbf{x}^{\prime}\,\Pr(\mathbf{x}^{\prime}|z)\,\left\Vert \mathbf{x}-\mathbf{x}^{\prime}\right\Vert ^{2}
\end{equation}
Now introduce dummy integrations over $y$ to obtain 
\begin{equation}
D_{\mathrm{VQ}}=\int d\mathbf{x}\,\Pr(\mathbf{x})\sum_{y=1}^{M_{y}}\Pr(y|\mathbf{x})\sum_{z=1}^{M_{z}}\Pr(z|y)\sum_{y^{\prime}=1}^{M_{y}}\Pr(y^{\prime}|z)\int d\mathbf{x}^{\prime}\,\Pr(\mathbf{x}^{\prime}|y^{\prime})\,\left\Vert \mathbf{x}-\mathbf{x}^{\prime}\right\Vert ^{2}
\end{equation}
and rearrange to obtain 
\begin{equation}
D_{\mathrm{VQ}}=\int d\mathbf{x}\,\Pr(\mathbf{x})\sum_{y^{\prime}=1}^{M_{y}}\Pr(y^{\prime}|\mathbf{x})\int d\mathbf{x}^{\prime}\,\Pr(\mathbf{x}^{\prime}|y^{\prime})\,\left\Vert \mathbf{x}-\mathbf{x}^{\prime}\right\Vert ^{2}
\end{equation}
where
\begin{eqnarray}
\Pr(y^{\prime}|y) & = & \sum_{z=1}^{M_{z}}\Pr(y^{\prime}|z)\Pr(z|y)\nonumber \\
\Pr(y^{\prime}|\mathbf{x}) & = & \sum_{y=1}^{M_{y}}\Pr(y^{\prime}|y)\Pr(y|\mathbf{x})\label{eq:LeakedPosterior}
\end{eqnarray}
which may be replaced by the equivalent objective function 
\begin{equation}
D_{\mathrm{VQ}}=2\int d\mathbf{x}\,\Pr(\mathbf{x})\sum_{y^{\prime}=1}^{M_{y}}\Pr(y^{\prime}|\mathbf{x})\,\left\Vert \mathbf{x}-\mathbf{x}^{\prime}(y^{\prime})\right\Vert ^{2}\label{eq:ObjectiveLayer01}
\end{equation}
By manipulating $D_{\mathrm{VQ}}$ from the form it has in equation
\ref{eq:ObjectiveLayer12} to the form it has in equation \ref{eq:ObjectiveLayer01},
it becomes clear that optimisation of the $(\mathbf{x},z)$ network
involves optimisation of the $(\mathbf{x},y^{\prime})$ subnetwork,
for which an objective function can be written that uses a $\Pr(y^{\prime}|\mathbf{x})$
as defined in equation \ref{eq:LeakedPosterior}. When optimising
the $(\mathbf{x},y^{\prime})$ subnetwork, $\Pr(y^{\prime}|y)$ takes
account of the effect that $z$ has on $y$. 

If $n=1$, so that only 1 firing event is observed, then $D_{\mathrm{VQ}}=D_{1}$,
and the optimum $\Pr(y|\mathbf{x})$ must ensure that $y$ depends
deterministically on $\mathbf{x}$, so that $\Pr(y|\mathbf{x})=\delta_{y,y(\mathbf{x})}$
where $y(\mathbf{x})$ is an encoding function that converts $\mathbf{x}$
into the index of the neuron that fires in response to $\mathbf{x}$.
This allows $D_{\mathrm{VQ}}$ to be simplified to 
\begin{equation}
D_{\mathrm{VQ}}=2\int d\mathbf{x}\,\Pr(\mathbf{x})\sum_{y^{\prime}=1}^{M_{y}}\Pr(y^{\prime}|y(\mathbf{x}))\,\left\Vert \mathbf{x}-\mathbf{x}^{\prime}(y^{\prime})\right\Vert ^{2}
\end{equation}
where $\Pr(y^{\prime}|y(\mathbf{x}))$ is $\Pr(y^{\prime}|y)$ with
$y$ replaced by $y(\mathbf{x})$. Note that if $\Pr(y^{\prime}|y)=\delta_{y,y^{\prime}}$
then $D_{\mathrm{VQ}}$ reduces to the objective function $2\int d\mathbf{x}\,\Pr(\mathbf{x})\,\left\Vert \mathbf{x}-\mathbf{x}^{\prime}(y(\mathbf{x}))\right\Vert ^{2}$
for a standard vector quantiser (VQ). 

The optimum $y(\mathbf{x})$ is given by $y(\mathbf{x})=\begin{array}{c}
\arg\min\\
y
\end{array}\sum_{y^{\prime}=1}^{M_{y}}\Pr(y^{\prime}|y)\,\left\Vert \mathbf{x}-\mathbf{x}^{\prime}(y^{\prime})\right\Vert ^{2}$ (which is not quite the same as the $y(\mathbf{x})=\begin{array}{c}
\arg\min\\
y
\end{array}\left\Vert \mathbf{x}-\mathbf{x}^{\prime}(y)\right\Vert ^{2}$ used by Kohonen in his topographic mapping neural network \cite{Kohonen1989}),
and a gradient descent algorithm for updating $\mathbf{x}^{\prime}(y^{\prime})$
is $\mathbf{x}^{\prime}(y^{\prime})\longrightarrow\mathbf{x}^{\prime}(y^{\prime})+\varepsilon\,\Pr(y^{\prime}|y(\mathbf{x}))$
(which is identical to Kohonen's prescription \cite{Kohonen1989}).
The $\Pr(y^{\prime}|y)$ may thus be interpreted as the neighbourhood
function, and the $\mathbf{x}^{\prime}(y^{\prime})$ may be interpreted
as the weight vectors, of a topographic mapping. Because all states
$y$ that can give rise to the same state $z$ (as specified by $\Pr(z|y)$)
become neighbours (as specified by $\Pr(y^{\prime}|y)$ in equation
\ref{eq:LeakedPosterior}), $\Pr(y^{\prime}|y)$ includes a much larger
class of neighbourhood functions than has hitherto been used in topographic
mapping neural networks.

Because of the principled way in which the topographic mapping objective
function has been derived here, it is the preferred way to optimise
topographic mapping networks. It also allows the objective function
to be generalised to the case $n>1$, where more than one firing event
is observed.

\subsection{Piecewise Linear Probability of Firing\label{sub:Piecewise-Linear-Probability}}

The optimal $\Pr(y|\mathbf{x})$ has some interesting properties that
can be obtained by inspecting its stationarity condition. For instance,
the $\Pr(y|\mathbf{x})$ that minimise $D_{1}+D_{2}$ will be shown
to be piecewise linear functions of $\mathbf{x}$. 

Thus, functionally differentiate $D_{1}+D_{2}$ with respect to $\log\Pr(y|\mathbf{x})$,
where logarithmic differentation implicitly imposes the constraint
$\Pr(y|\mathbf{x})\geq0$, and use a Lagrange multiplier term $L\equiv\int d\mathbf{x}^{\prime}\,\lambda(\mathbf{x}^{\prime})\,\sum_{y^{\prime}=1}^{M}\Pr(y^{\prime}|\mathbf{x}^{\prime})$
to impose the normalisation constraint $\sum_{y=1}^{M}\Pr(y|\mathbf{x})=1$
for each $\mathbf{x}$, to obtain 
\begin{eqnarray}
\frac{\delta\left(D_{1}+D_{2}-L\right)}{\delta\log\Pr\left(y|\mathbf{x}\right)} & = & \frac{2}{n}\,\Pr\left(\mathbf{x}\right)\,\Pr(y|\mathbf{x})\left\Vert \mathbf{x}-\mathbf{x}^{\prime}(y)\right\Vert ^{2}\nonumber \\
 &  & -\frac{4(n-1)}{n}\,\Pr(\mathbf{x})\,\Pr(y|\mathbf{x})\,\mathbf{x}^{\prime}(y)\,\cdot\left(\mathbf{x}-\sum_{y=1}^{M}\Pr(y|\mathbf{x})\,\mathbf{x}^{\prime}(y)\right)\nonumber \\
 &  & -\lambda(\mathbf{x})\,\Pr(y|\mathbf{x})
\end{eqnarray}
The stationarity condition implies that $\sum_{y=1}^{M}\Pr(y|\mathbf{x})\,\frac{\delta\left(D_{1}+D_{2}-L\right)}{\delta\Pr(y|\mathbf{x})}=0$,
which may be used to determine the Lagrange multiplier function $\lambda(\mathbf{x})$.
When $\lambda(\mathbf{x})$ is substituted back into the stationarity
condition itself, it yields 
\begin{multline}
0=\Pr(\mathbf{x})\,\Pr(y|\mathbf{x})\sum_{y^{\prime}=1}^{M}\left(\Pr(y^{\prime}|\mathbf{x})-\delta_{y,y^{\prime}}\right)\\
\times\mathbf{x}^{\prime}(y^{\prime})\,\cdot\left(\frac{\mathbf{x}^{\prime}(y^{\prime})}{2}-n\,\mathbf{x+}(n-1)\sum_{y^{\prime\prime}=1}^{M}\Pr(y^{\prime\prime}|\mathbf{x})\,\mathbf{x}^{\prime}(y^{\prime\prime})\right)\label{eq:StationarityP}
\end{multline}
There are several classes of solution to this stationarity condition,
corresponding to one (or more) of the three factors in equation \ref{eq:StationarityP}
being zero. 
\begin{enumerate}
\item $\Pr(\mathbf{x})=0$ (the first factor is zero). If the input PDF
is zero at $\mathbf{x}$, then nothing can be deduced about $\Pr(y|\mathbf{x})$,
because there is no training data to explore the network's behaviour
at this point. 
\item $\Pr(y|\mathbf{x})=0$ (the second factor is zero). This factor arises
from the differentiation with respect to $\log\Pr(y|\mathbf{x})$,
and it ensures that $\Pr(y|\mathbf{x})<0$ cannot be attained. The
singularity in $\log\Pr(y|\mathbf{x})$ when $\Pr(y|\mathbf{x})=0$
is what causes this solution to emerge. 
\item $\sum_{y^{\prime}=1}^{M}\left(\Pr(y^{\prime}|\mathbf{x})-\delta_{y,y^{\prime}}\right)\,\mathbf{x}^{\prime}(y^{\prime})\cdot(\cdots)=0$
(the third factor is zero). The solution to this equation is a $\Pr(y|\mathbf{x})$
that has a piecewise linear dependence on $\mathbf{x}$. This result
can be seen to be intuitively reasonable because $D_{1}+D_{2}$ is
of the form $\int d\mathbf{x}\,\Pr(\mathbf{x})\,f(\mathbf{x})$, where
$f(\mathbf{x})$ is a linear combination of terms of the form $\mathbf{x}^{i}\,\Pr(y|\mathbf{x})^{j}$
(for $i=0,1,2$ and $j=0,1,2$), which is a quadratic form in $\mathbf{x}$
(ignoring the $\mathbf{x}$-dependence of $\Pr(y|\mathbf{x})$). However,
the terms that appear in this linear combination are such that a $\Pr(y|\mathbf{x})$
that is a piecewise linear function of $\mathbf{x}$ guarantees that
$f(\mathbf{x})$ is a piecewise linear combination of terms of the
form $\mathbf{x}^{i}$ (for $i=0,1,2$), which is a quadratic form
in $\mathbf{x}$ (the normalisation constraint $\sum_{y=1}^{M}\Pr(y|\mathbf{x})=1$
is used to remove a contribution to that is potentially quartic in
$\mathbf{x}$). Thus a piecewise linear dependence of $\Pr(y|\mathbf{x})$
on $\mathbf{x}$ does not lead to any dependencies on $\mathbf{x}$
that are not already explicitly present in $D_{1}+D_{2}$. The stationarity
condition on $\Pr(y|\mathbf{x})$ (see equation \ref{eq:StationarityP})
then imposes conditions on the allowed piecewise linearities that
$\Pr(y|\mathbf{x})$ can have.
\end{enumerate}
For the purpose of doing analytic calculations, it is much easier
to obtain analytic results with the ideal piecewise linear $\Pr(y|\mathbf{x})$
than with some other functional form. If the optimisation of $\Pr(y|\mathbf{x})$
is constrained, by introducing a parametric form which has some biological
plausibility, for instance, then analytic optimum solutions are not
in general possible to calculate, and it becomes necessary to resort
to numerical simulations. Piecewise linear $\Pr(y|\mathbf{x})$ should
therefore be regarded as a convenient theoretical laboratory for investigating
the properties of idealised neural networks.

\subsubsection{Solved Example}

A simple example illustrates how the piecewise linearity property
of $\Pr(y|\mathbf{x})$ may be used to find optimal solutions. Thus
consider a 1-dimensional input coordinate $x\in[-\infty,+\infty]$,
with $\Pr(x)=P_{0}$. Assume that the number of neurons $M$ tends
to infinity in such a way that there is 1 neuron per unit length of
$x$, so that $\Pr(y|x)=p(y-x)$, where the piecewise linear property
gives $p(x)$ as 
\begin{equation}
p(x)=\left\{ \begin{array}{lllll}
1 &  &  &  & \left|x\right|\leq\frac{1}{2}-s\\
\frac{2s-2\left|x\right|+1}{4s} &  &  &  & \frac{1}{2}-s\leq\left|x\right|\leq\frac{1}{2}+s\\
0 &  &  &  & \left|x\right|\geq\frac{1}{2}+s
\end{array}\right.
\end{equation}
and by symmetry $x^{\prime}(y)=y$.

This $\Pr(y|x)$ and $x^{\prime}(y)$ allow $D_{1}$ to be derived
as 
\begin{eqnarray}
D_{1}\;(\text{per neuron}) & = & \frac{2P_{0}}{n}\left(\int_{-\frac{1}{2}+s}^{\frac{1}{2}-s}dx\,x^{2}+2\int_{\frac{1}{2}-s}^{\frac{1}{2}+s}dx\,\frac{2s-2x+1}{4s}\,x^{2}\right)\nonumber \\
 & = & \frac{P_{0}}{6n}\,\left(1+4s^{2}\right)
\end{eqnarray}
and $D_{2}$ to be derived as

\begin{eqnarray}
D_{2}\;\left(\text{per unit length}\right) & = & \frac{2\left(n-1\right)P_{0}}{n}\left(\begin{array}{l}
\int_{-\frac{1}{2}+s}^{\frac{1}{2}-s}x^{2}\,dx\\
\\
+2\int_{\frac{1}{2}-s}^{\frac{1}{2}}\left(x-\frac{2s+2\left(x-1\right)+1}{4s}\right)^{2}\,dx
\end{array}\right)\nonumber \\
 & = & \frac{\left(n-1\right)P_{0}}{6n}\,\left(2s-1\right)^{2}
\end{eqnarray}
Because there is one neuron per unit length, the contribution per
unit length to $D_{1}+D_{2}$ is the sum of the above two results
\begin{equation}
D_{1}+D_{2}\;(\text{per unit length})=\frac{P_{0}}{6n}\,\left(n\left(2s-1\right)^{2}+4s\right)
\end{equation}
If $D_{1}+D_{2}$ is differentiated with respect to $s$, then stationarity
condition $\frac{d\left(D_{1}+D_{2}\right)}{ds}=0$ yields the optimum
value of $s$ as 
\begin{equation}
s=\frac{n-1}{2n}
\end{equation}
and the stationary value of $D_{1}+D_{2}$ as 
\begin{equation}
D_{1}+D_{2}\;(\text{per unit length})=\frac{\left(2n-1\right)\,P_{0}}{6n^{2}}
\end{equation}

When $n=1$ the stationary solution reduces to $s=0$ and $D_{1}+D_{2}$
(per unit length) $=\frac{P_{0}}{6}$, which is a standard vector
quantiser with nonoverlapping neural response regions which partition
the input space into unit width quantisation cells, so that for all
$x$ there is exactly one neuron that responds. Although the neurons
have been manually arranged in topographic order by imposing $\Pr(y|x)=p(y-x)$,
any permutation of the neuron indices in this stationary solution
will also be stationary solution. This derivation could be generalised
to the type of 3-layer network that was considered in section \ref{sub:Topographic-Mapping-Neural}
, in which case a neighbourhod function $\Pr(y^{\prime}|y)$ would
emerge automatically. 

As $n\longrightarrow\infty$ the stationary solution behaves as $s\longrightarrow\frac{1}{2}$
and $D_{1}+D_{2}$ (per unit length) $\longrightarrow\frac{P_{0}}{3n}$,
with overlapping linear neural response regions which cover the input
space, so that for all $x$ there are exactly two neurons that respond
with equal and opposite linear dependence on $x$. As $n\longrightarrow\infty$
the ratio of the number of firing events that occur on these two neurons
is sufficient to determine $x$ to ${\cal O}\left(\frac{1}{n}\right)$.
When $n=\infty$ this stationary solution is $s=\frac{1}{2}$ and
$D_{1}+D_{2}$ (per unit length) $=0$. However, when $n=\infty$
there are infinitely many other ways in which the neurons could be
used to yield $D_{1}+D_{2}$ (per unit length) $=0$, because only
the $D_{2}$ term contributes, and it is 0 when $x=\sum_{y=1}^{M}\Pr(y|x)\,x^{\prime}(y)$.
This is possible for any set of basis elements $x^{\prime}(y)$ that
span the input space, provided that the expansion coefficients $\Pr(y|x)$
satisfy $\Pr(y|x)\geq0$. In this 1-dimensional example only two basis
elements are required (i.e. $M=2$), which are $x^{\prime}(1)=-\infty$
and $x^{\prime}(2)=+\infty$. More generally, for this type of stationary
solution, $M=\dim\mathbf{x}+1$ is required to span the input space
in such a way that $\Pr(y|x)\geq0$, and if $M<\dim\mathbf{x}+1$
then the stationary solution will span the input subspace (of dimension
$M-1$) that has the largest variance. 

The $n=1$ and $n\longrightarrow\infty$ limiting cases are very different.
When $n=1$ the optimum network splits up the input space into non-overlapping
quantisation cells, and as $n\longrightarrow\infty$ the optimum network
does a linear decomposition of the input space using non-negative
expansion coefficients. This behaviour occurs because for $n>1$ the
neurons can cooperate when encoding the input $x$, so that by allowing
more than one neuron to fire in response to $x$, the encoded version
of $x$ is distributed over more than one neuron. In the above 1-dimensional
example, the code is spread over one or two neurons depending on the
value of $x$. This cooperation amongst neurons is a property of the
coherent part $D_{2}$ of the upper bound on $D_{\mathrm{VQ}}$ (see
equation \ref{eq:ObjectiveD1D2}).

\subsection{Factorial Encoder Network\label{sub:Factorial-Encoder-Network}}

For certain types of distribution of data in input space the optimal
network consists of a number of subnetworks, each of which responds
to only a subspace of the input space. This is called factorial encoding,
where the encoded input is distributed over more than one neuron,
and this distributed code typically has a much richer structure than
was encountered in section \ref{sub:Piecewise-Linear-Probability}. 

The simplest problem that demonstrates factorial encoding will now
be investigated (this example was presented in \cite{Luttrell1997b},
but the derivation given here is more direct). Thus, assume that the
data in input space  uniformly populates the surface of a 2-torus
$S^{1}\times S^{1}$. Each of the $S^{1}$ is a plane unit circle
embedded in $R^{1}\times R^{1}$ and centred on the origin, and $S^{1}\times S^{1}$
is the Cartesian product of a pair of such circles. Overall, the 2-torus
lives in a 4-dimensional input space whose elements are denoted as
$\mathbf{x}=(x_{1},x_{2},x_{3},x_{4})$, where one of the circles
lives in $(x_{1},x_{2})$ and the other lives in $(x_{3},x_{4})$.
These circles may be parameterised by angular degrees of freedom $\theta_{12}$
and $\theta_{34}$, respectively. 

The optimal $\Pr(y|\mathbf{x})$ (i.e. a piecewise linear stationary
solution of the type that was encountered in section \ref{sub:Piecewise-Linear-Probability}
could be derived from this input data PDF $\Pr(\mathbf{x})$. However,
the properties of the sought-after optimal $\Pr(y|\mathbf{x})$ are
preserved if one restricts the solution space to the following types
of $\Pr(y|\mathbf{x})$ 
\begin{equation}
\Pr(y|\mathbf{x})=\left\{ \begin{array}{lllll}
\delta_{y,y(\theta_{12})}\text{ or }\delta_{y,y(\theta_{34})} &  &  &  & \text{type 1}\\
\\
\delta_{y,y(\theta_{12},\theta_{34})} &  &  &  & \text{type 2}\\
\\
\frac{1}{2}\left(\delta_{y,y_{12}(\theta_{12})}+\delta_{y,y_{34}(\theta_{34})}\right) &  &  &  & \text{type 3}
\end{array}\right.\label{eq:SolutionTypes}
\end{equation}
where $y(\theta_{12})$ and $y_{12}(\theta_{12})$ encode $\theta_{12}$,
$y(\theta_{34})$ and $y_{34}(\theta_{34})$ encode $\theta_{34}$,
and $y(\theta_{12},\theta_{34})$ encodes $(\theta_{12},\theta_{34})$.
The allowed ranges of the code indices are $1\leq y(\theta_{12})\leq M$
(and similarly $y(\theta_{34})$), $1\leq y_{12}(\theta_{12})\leq\frac{M}{2}$,
$\frac{M}{2}+1\leq y_{34}(\theta_{34})\leq M$, and $1\leq y(\theta_{12},\theta_{34})\leq M$.
The type 1 solution assumes that all $M$ neurons respond only to
$\theta_{12}$ (or, alternatively, all respond only to $\theta_{34}$),
the type 2 solution assumes that all $M$ neurons respond to $(\theta_{12},\theta_{34})$,
and the type 3 solution (which is very simple type of factorial encoder)
assumes that $\frac{M}{2}$ neurons respond only to $\theta_{12}$,
and the other $\frac{M}{2}$ neurons respond only to $\theta_{34}$. 

In order to derive explicit results for the stationary value of $D_{1}+D_{2}$,
it is necessary to optimise the $\mathbf{x}^{\prime}(y)$. The stationary
condition on $\mathbf{x}^{\prime}(y)$ may readily be deduced from
the stationarity condition $\frac{\partial(D_{1}+D_{2})}{\partial\mathbf{x}^{\prime}(y)}=0$
as 
\begin{equation}
n\int d\mathbf{x}\,\Pr(\mathbf{x}|y)\,\mathbf{x=x}^{\prime}(y)+(n-1)\int d\mathbf{x}\,\Pr(\mathbf{x}|y)\sum_{y^{\prime}=1}^{M}\Pr(y^{\prime}|\mathbf{x})\,\mathbf{x}^{\prime}(y^{\prime})
\end{equation}
If $\Pr(y|\mathbf{x})$ (and hence $\Pr(\mathbf{x}|y)$) are inserted
into this stationarity condition, then it may be solved for the corresponding
$\mathbf{x}^{\prime}(y)$.

Assume that the encoding functions partition up the 2-torus symmetrically,
the three types of solution may be optimised as described in the following
three sections.

\subsubsection{Type 1 Solution}

Assume that $\Pr(y|\mathbf{x})=\delta_{y,y(x_{1},x_{2})}$, and that
the $y=1$ quantisation cell is the Cartesian product of the arcs
$\left|\theta_{12}\right|\leq\frac{\pi}{M}$ and $\theta_{34}\leq2\pi$
of the 2 unit circles that form the 2-torus, then the stationarity
condition for $\mathbf{x}^{\prime}(1)$ becomes

\begin{multline}
n\,\frac{M}{2\pi}\int_{-\frac{\pi}{M}}^{\frac{\pi}{M}}d\theta_{12}\,\frac{1}{2\pi}\int_{0}^{2\pi}d\theta_{34}\,\left(\cos\theta_{12},\sin\theta_{12},\cos\theta_{34},\sin\theta_{34}\right)\\
=\mathbf{x}^{\prime}(1)+(n-1)\,\frac{M}{2\pi}\int_{-\frac{\pi}{M}}^{\frac{\pi}{M}}d\theta_{12}\,\frac{1}{2\pi}\int_{0}^{2\pi}d\theta_{34}\,\mathbf{x}^{\prime}(1)
\end{multline}
which yields the solution $\mathbf{x}^{\prime}(1)=\left(\frac{M}{\pi}\,\sin\left(\frac{\pi}{M}\right),0,0,0\right)$.
The first two components are the centroid of the arc $\left|\vartheta\right|\leq\frac{\pi}{M}$
of a unit circle centred on the origin. All of the $\mathbf{x}^{\prime}(y)$
can be obtained by rotating $\mathbf{x}^{\prime}(1)$ about the origin
by multiples of $\frac{2\pi}{M}$. Using the assumed symmetry of the
solution, the expression for $D_{1}+D_{2}$ becomes
\begin{multline}
D_{1}+D_{2}=\frac{M}{\pi}\int_{-\frac{\pi}{M}}^{\frac{\pi}{M}}d\theta_{12}\,\left\Vert \left(\cos\theta_{12},\sin\theta_{12}\right)-\left(\frac{M}{\pi}\,\sin\left(\frac{\pi}{M}\right),0\right)\right\Vert ^{2}\\
+\frac{1}{\pi}\int_{0}^{2\pi}d\theta_{34}\,\left\Vert \left(\cos\theta_{34},\sin\theta_{34}\right)-\left(0,0\right)\right\Vert ^{2}\label{eq:D1D2Type1}
\end{multline}
where the first (or second) term corresponds to the subspace to which
the neurons respond (or not respond). This gives the stationary value
of $D_{1}+D_{2}$ as $D_{1}+D_{2}=4-\frac{2M^{2}}{\pi^{2}}\,\sin^{2}\left(\frac{\pi}{M}\right)$.
Only one neuron can fire (given $\mathbf{x}$), because $\Pr(y|\mathbf{x})=\delta_{y,y(\theta_{12})}$
or $\delta_{y,y(\theta_{34})}$, no further information about $\mathbf{x}$
can be obtained after the first firing event has occurred, so this
result for $D_{1}+D_{2}$ is independent of $n$, as expected.

\subsubsection{Type 2 Solution}

Assume that the $y=1$ quantisation cell is the Cartesian product
of the arcs $\left|\vartheta_{12}\right|\leq\frac{\pi}{\sqrt{M}}$
and $\left|\vartheta_{34}\right|\leq\frac{\pi}{\sqrt{M}}$ of the
two unit circles that form the 2-torus. The stationarity condition
for $\mathbf{x}^{\prime}(1)$ can be deduced from the type 1 case
with the replacement $M\longrightarrow\sqrt{M}$, which gives $\mathbf{x}^{\prime}(1)=\left(\frac{\sqrt{M}}{\pi}\,\sin\left(\frac{\pi}{\sqrt{M}}\right),0,\frac{\sqrt{M}}{\pi}\,\sin\left(\frac{\pi}{\sqrt{M}}\right),0\right)$.
The expression for $D_{1}+D_{2}$ may similarly be deduced from the
type 1 case as twice the first term in equation \ref{eq:D1D2Type1}
with the replacement $M\longrightarrow\sqrt{M}$, to yield the stationary
value of $D_{1}+D_{2}$ as $D_{1}+D_{2}=4-\frac{4M}{\pi^{2}}\,\sin^{2}\left(\frac{\pi}{\sqrt{M}}\right)$.
As in the type 1 case, this result for $D_{1}+D_{2}$ is independent
of $n$.

\subsubsection{Type 3 Solution}

The stationarity condition for $\mathbf{x}^{\prime}(1)$ can be written
by analogy with the type 1 case, with the replacement $M\longrightarrow\frac{M}{2}$,
and modifying the last term to take account of the more complicated
form of $\Pr(y|\mathbf{x})$, to yield
\begin{multline}
n\,\frac{M}{4\pi}\int_{-\frac{2\pi}{M}}^{\frac{2\pi}{M}}d\theta_{12}\,\frac{1}{2\pi}\int_{0}^{2\pi}d\theta_{34}\,\left(\cos\theta_{12},\sin\theta_{12},\cos\theta_{34},\sin\theta_{34}\right)\\
=\mathbf{x}^{\prime}(1)+\frac{1}{2}(n-1)\,\frac{M}{4\pi}\int_{-\frac{2\pi}{M}}^{\frac{2\pi}{M}}d\theta_{12}\,\frac{1}{2\pi}\int_{0}^{2\pi}d\theta_{34}\,\mathbf{x}^{\prime}(1)
\end{multline}
where $\frac{1}{2\pi}\int_{0}^{2\pi}d\theta_{34}\sum_{y^{\prime}=\frac{M}{2}+1}^{M}\delta_{y^{\prime},y_{34}(\theta_{34})}\,\mathbf{x}^{\prime}(y^{\prime})=0$
has been used (this follows from the assumed symmetry of the solution).
This yields the solution $\mathbf{x}^{\prime}(1)=\frac{2n}{n+1}\left(\frac{M}{2\pi}\,\sin\left(\frac{2\pi}{M}\right),0,0,0\right)$.
Using the assumed symmetry of the solution, the expression for $D_{1}$
becomes

\begin{multline}
D_{1}=\frac{2}{n}\left(\frac{M}{4\pi}\int_{-\frac{2\pi}{M}}^{\frac{2\pi}{M}}d\theta_{12}\,\left\Vert \left(\cos\theta_{12},\sin\theta_{12}\right)-\frac{2n}{n+1}\left(\frac{M}{2\pi}\,\sin\left(\frac{2\pi}{M}\right),0\right)\right\Vert ^{2}\right.\\
+\left.\frac{1}{2\pi}\int_{0}^{2\pi}d\theta_{34}\left\Vert \left(\cos\theta_{34},\sin\theta_{34}\right)\right\Vert ^{2}\vphantom{\left\Vert \sin\left(\frac{2\pi}{M}\right)\right\Vert ^{2}}\right)
\end{multline}
and the expression for $D_{2}$ becomes 
\begin{equation}
D_{2}=\frac{2(n-1)}{n}\left(\frac{M}{2\pi}\int_{-\frac{2\pi}{M}}^{\frac{2\pi}{M}}\,d\theta_{12}\left\Vert \left(\cos\theta_{12},\sin\theta_{12}\right)-\frac{n}{n+1}\left(\frac{M}{2\pi}\,\sin\left(\frac{2\pi}{M}\right),0\right)\right\Vert ^{2}\right)
\end{equation}
This gives the stationary value of $D_{1}+D_{2}$ as $D_{1}+D_{2}=4-\frac{nM^{2}}{(n+1)\pi^{2}}\,\sin^{2}\left(\frac{2\pi}{M}\right)$.
Because $\Pr(y|\mathbf{x})=\frac{1}{2}\left(\delta_{y,y_{12}(\theta_{12})}+\delta_{y,y_{34}(\theta_{34})}\right)$,
one firing event has to occur in each of the intervals $1\leq y\leq\frac{M}{2}$
and $\frac{M}{2}+1\leq y\leq M$ for all of the information to be
collected about $\mathbf{x}$. However, the random nature of the firing
events means that the probability with which this condition is satisfied
increases with $n$, so this result for $D_{1}+D_{2}$ decreases as
$n$ increases.

\subsubsection{Relative Stability of Solutions}

Collect the above results together for comparison. 
\begin{equation}
D_{1}+D_{2}=\left\{ \begin{array}{lllll}
4-\frac{2M^{2}}{\pi^{2}}\,\sin^{2}\left(\frac{\pi}{M}\right) &  &  &  & \text{type 1}\\
\\
4-\frac{4M}{\pi^{2}}\,\sin^{2}\left(\frac{\pi}{\sqrt{M}}\right) &  &  &  & \text{type 2}\\
\\
4-\frac{nM^{2}}{(n+1)\pi^{2}}\,\sin^{2}\left(\frac{2\pi}{M}\right) &  &  &  & \text{type 3}
\end{array}\right.
\end{equation}
For constant $M$ and letting $n\longrightarrow\infty$, the value
of $D_{1}+D_{2}$ for the type 3 solution asymptotically behaves as
$D_{1}+D_{2}\;(\text{type 3})\longrightarrow4-\frac{M^{2}}{\pi^{2}}\,\sin^{2}\left(\frac{2\pi}{M}\right)$,
in which case the relative stability of the three types of solution
is: type 3 (most stable), type 2 (intermediate), type 1 (least stable).
Similarly, for constant $n$ and letting $M\longrightarrow\infty$,
the relative stability of the three types of solution is: type 2 (most
stable), type 3 (intermediate), type 1 (least stable). 

In both of these limiting cases the type 1 solution is least stable.
If there is a fixed number of firing events $n$, and there is no
upper limit on the number of neurons $M$, then the type 2 solution
is most stable, because it can partition the 2-torus into lots of
small quantisation cells. If there is a fixed number of neurons $M$
(which is the usual case), and there is no upper limit on the number
of firing events $n$, then the type 3 solution is most stable, because
the limited size of $M$ renders the type 2 solution inefficient (the
quantisation cells would be too large), so the 2-torus $S^{1}\times S^{1}$
is split into two $S^{1}$ subspaces each of which is assigned a subset
of $\frac{M}{2}$ neurons. If $n$ is large enough, then each of these
two subsets of neurons has a high probability of occurrence of a firing
event, which ensures that both of the $S^{1}$ subspaces are encoded.

More generally, when there is a limited number of neurons they will
tend to split into subsets, each of which encodes a separate subspace
of the input. The assumed form of $\Pr(y|\mathbf{x})$ in equation
\ref{eq:SolutionTypes} does not allow an unrestricted search of all
possible $\Pr(y|\mathbf{x})$. If the global optimum solution (which
has piecewise linear $\Pr(y|\mathbf{x})$, as proved in section \ref{sub:Piecewise-Linear-Probability})
cuts up the input space into partially overlapping pieces, then it
is well approximated by a solution such as one of those listed in
equation \ref{eq:SolutionTypes}. Typically, curved input spaces lead
to such solutions, because a piecewise linear $\Pr(y|\mathbf{x})$
can readily quantise such spaces by slicing off the curved ``corners'''
that occur in such spaces.

\section{Conclusions}

In this paper an objective function for optimising a layered network
of discretely firing neurons has been presented, and three non-trivial
examples of how it is applied have been shown: topographic mapping
networks, piecewise linear dependence on the input of the probability
of a neuron firing, and factorial encoder networks. Many other examples
could be given, such as combining the first and third of the above
results to obtain factorial topographic networks, or extending the
theory to multilayer networks, or introducing temporal information.

\section{Acknowledgements}

I thank Chris Webber for many useful conversations that we had during
the course of this research.

\bibliographystyle{plain}
\bibliography{../../biblio}

\begin{thebibliography}{1}

\bibitem{Kohonen1989}
T~Kohonen.
\newblock {\em Self-organisation and associative memory}.
\newblock Springer-Verlag, Berlin, 1989.

\bibitem{Luttrell1989b}
S~P Luttrell.
\newblock Hierarchical self-organising networks.
\newblock In {\em Proceedings of IEE Conference on Artificial Neural Networks},
  pages 2--6, London, 1989. IEE.

\bibitem{Luttrell1997}
S~P Luttrell.
\newblock {\em Mathematics of neural networks: models, algorithms and
  applications}, chapter A theory of self-organising neural networks, pages
  240--244.
\newblock Kluwer, Boston, 1997.

\bibitem{Luttrell1997b}
S~P Luttrell.
\newblock Self-organisation of multiple winner-take-all neural networks.
\newblock {\em Connection Science}, 9(1):11--30, 1997.

\bibitem{Rissanen1978}
J~Rissanen.
\newblock Modelling by shortest data description.
\newblock {\em Automatica}, 14(5):465--471, 1978.

\end{thebibliography}

\end{document}